\documentclass[12pt]{article}

\usepackage{scicite}
\usepackage{graphicx}
\usepackage{times}
\usepackage[dvipsnames]{xcolor}
\usepackage{hyperref}

\hypersetup{
    colorlinks=true,
    linkcolor=blue,
    filecolor=magenta,      
    urlcolor=cyan,
    citecolor=blue,
    pdfpagemode=FullScreen,
    }

\newenvironment{sciabstract}{%
\begin{quote} \bf}
{\end{quote}}

\title{Video-rate multispectral imaging in laparoscopic surgery: First-in-human application}

\author
{Leonardo Ayala,$^{1,5\ast}$ 
Sebastian Wirkert,$^1$
Anant Vemuri,$^1$
Tim Adler,$^{1,2}$ \\
Silvia Seidlitz,$^{1,3}$
Sebastian Pirmann,$^1$
Christina Engels,$^4$ \\
Dogu Teber,$^4$ 
Lena Maier-Hein$^{1,2,5\ast\ast}$\\
\\
\normalsize{$^1$Division of Computer Assisted Medical Interventions,}\\ \normalsize{German Cancer Research Center (DKFZ), Heidelberg, Germany}\\
\normalsize{$^2$Faculty of Mathematics and Computer Science, Heidelberg University, Heidelberg, Germany}\\
\normalsize{$^3$HIDSS4Health - Helmholtz Information and Data Science School for Health,}\\ \normalsize{Karlsruhe/Heidelberg, Germany}\\
\normalsize{$^4$St\"adtisches Klinikum Karlsruhe, Karlsruhe, Germany}\\
\normalsize{$^5$Medical Faculty, Heidelberg University, Heidelberg, Germany}\\
\\
\normalsize{$^{\ast}$e-mail:l.menjivar@dkfz.de}\\
\normalsize{$^{\ast \ast}$e-mail:l.maier-hein@dkfz.de}
}

\date{}

\usepackage[
backend=bibtex,
style=alphabetic,
sorting=ynt
]{biblatex}

\providecommand{\keywords}[1]
{
  \small	
  \textbf{\textit{Keywords--}} #1
}

\bibliography{main}

\begin{document} 


\baselineskip24pt


\maketitle


\begin{sciabstract}
  Multispectral and hyperspectral imaging (MSI/HSI) can provide clinically relevant information on morphological and functional tissue properties. Application in the operating room (OR), however, has so far been limited by complex hardware setups and slow acquisition times. To overcome these limitations, we propose a novel imaging system for video-rate spectral imaging in the clinical workflow. The system integrates a small snapshot multispectral camera with a standard laparoscope and a clinically commonly used light source, enabling the recording of multispectral images with a spectral dimension of 16 at a frame rate of 25 Hz. An ongoing \textit{in patient} study shows that multispectral recordings from this system can help detect perfusion changes in partial nephrectomy surgery, thus opening the doors to a wide range of clinical applications. 
\end{sciabstract}

\keywords{Multispectral imaging, hyperspectral imaging, minimally-invasive surgery, surgical endoscopy, oxygenation estimation, tissue, classification, video-rate, real time}

\section{Introduction}

Minimal invasive procedures are often preferred over open surgeries because they result in smaller scars, fewer complications and a quicker recovery of the patients. However, they come at the cost of reduced mobility and perception limitations of the surgeon. In many laparoscopic surgeries, for example, it is necessary to stop the blood flow to a specific organ or tissue region by clamping the arteries responsible for blood supply. This procedure, commonly referred to as \textit{ischemia induction}, prevents excessive bleeding of patients~\cite{thompson_impact_2007}  and is performed in various procedures, including partial nephrectomy, organ transplantation and anastomosis. After clamping the main arteries, it is highly challenging to assess the perfusion state of the tissue solely based on the available RGB video stream. This holds especially true when selective clamping of a segmental artery is performed, in which ischemia is induced only in the cancerous part of the kidney during partial nephrectomy \cite{mcclintock_can_2014, borofsky_near-infrared_2013}

Traditional approaches to improving surgical vision involve fusing preoperatively acquired images with the situs~\cite{Maier-Hein2018}. Such "offline" methods, however, cannot react on dynamics. The most common approach to ensure correct clamping is based on indocyanine green (ICG) fluorescence: after ICG is injected into the blood stream, it binds to the plasma. The bound ICG travels through the blood stream and accumulates in the internal organs, especially in the kidney and liver, within a minute \cite{tobis_near_2012, gandaglia_novel_2016}. Lack of a fluorescent signal thus corresponds to lack of perfusion. Due to long washout periods of about 30 minutes, this test is not repeatable if the wrong segment has been clamped \cite{gandaglia_novel_2016}. 

An alternative approach to improving surgical vision is spectral imaging. Minimally-invasive procedures are typically performed with a laparoscope featuring a traditional RGB camera. These RGB cameras try to mimic the human eye by collecting light in three broad regions of the optical spectrum (red, green and blue), normally referred to as bands. In contrast, multispectral imaging (MSI) or hyperspectral imaging (HSI) aims to increase the number of bands by collecting light in narrower regions, which can generally extend beyond the optical spectrum \cite{Clancy2020,Li2013}. The potential of this method has already been shown in the context of perfusion monitoring by means of \textit{in vitro} experiments and porcine trials \cite{Wirkert2017,Ayala_2019,Wirkert2018_1000086188}. 

Despite the general successes of MSI and HSI~\cite{Clancy2020}, application in the operating room (OR) has been limited. One of the main reasons why MSI has not yet found its way into a practical surgical application is related to the image acquisition, processing time and the size of the devices required~\cite{Clancy2020}. In fact, many available MSI/HSI cameras require big devices or a long time to record and further process one image \cite{DIETRICH2021,Kohler2020,Hu2020,Takamatsu_2021}. This is problematic, especially in laparoscopic surgery where organ motion is very pronounced and can lead to image blurring, thus hindering translation of this technology into the clinics.

To address this gap in the literature, we present a novel video-rate multispectral system that is suited for minimally-invasive surgeries featuring an acquisition speed above 25 frames per second while being lightweight and small in size. We further showcase this multispectral system in laparoscopic partial nephrectomy and demonstrate how it can be used to monitor the perfusion state of the kidney.

\section{Material and Methods}
This section introduces our multispectral imaging system as well as an in-human proof-of-concept study investigating the suitability of the system for \textit{in vivo} perfusion monitoring.

\subsection{Multispectral imaging system}
\label{subsec:multispectral-system}
\begin{figure}[h!]
    \centering
    \includegraphics[width=0.9\textwidth]{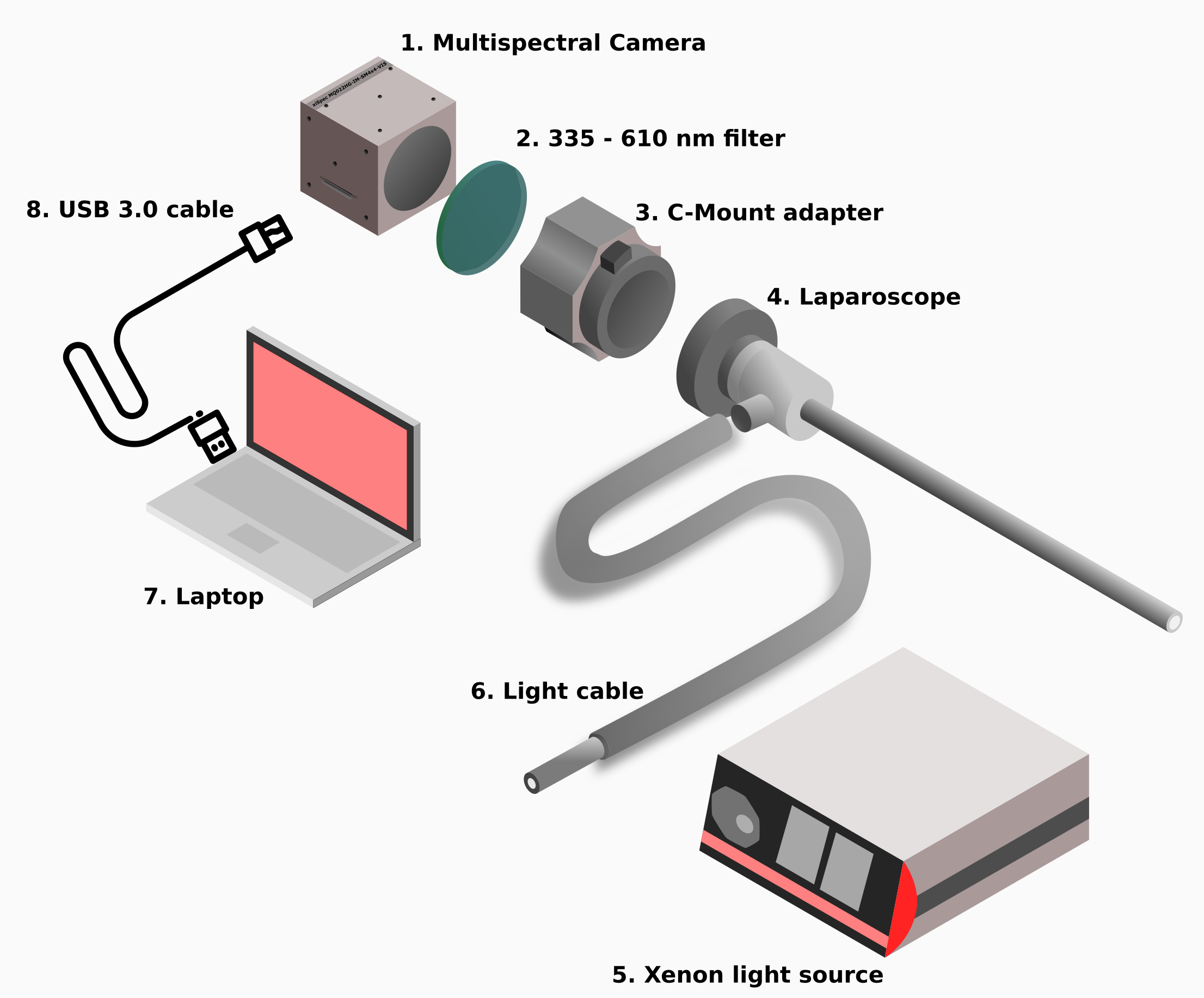}
    \caption{Schematic representation of the system developed for multispectral tissue analysis in laparoscopic surgery. It comprises (1) a snapshot multispectral camera, (2) a 335 - 610 nm band pass filter, (3) a C-Mount adapter with adjustable focal length, (4) a standard surgical laparoscope, (5) a Xenon light source, (6) a surgical light cable, (7) a laptop to control the multispectral camera and (8) a USB cable to connect the multispectral camera to the laptop. The dimensions of the laptop and the light source have been scaled down for visualization purposes.}
    \label{fig:msi_camera}
\end{figure}

Our imaging system is illustrated in Fig. \ref{fig:msi_camera} and comprises the following main components.

\begin{figure}
    \centering
    \includegraphics[width=0.6\textwidth]{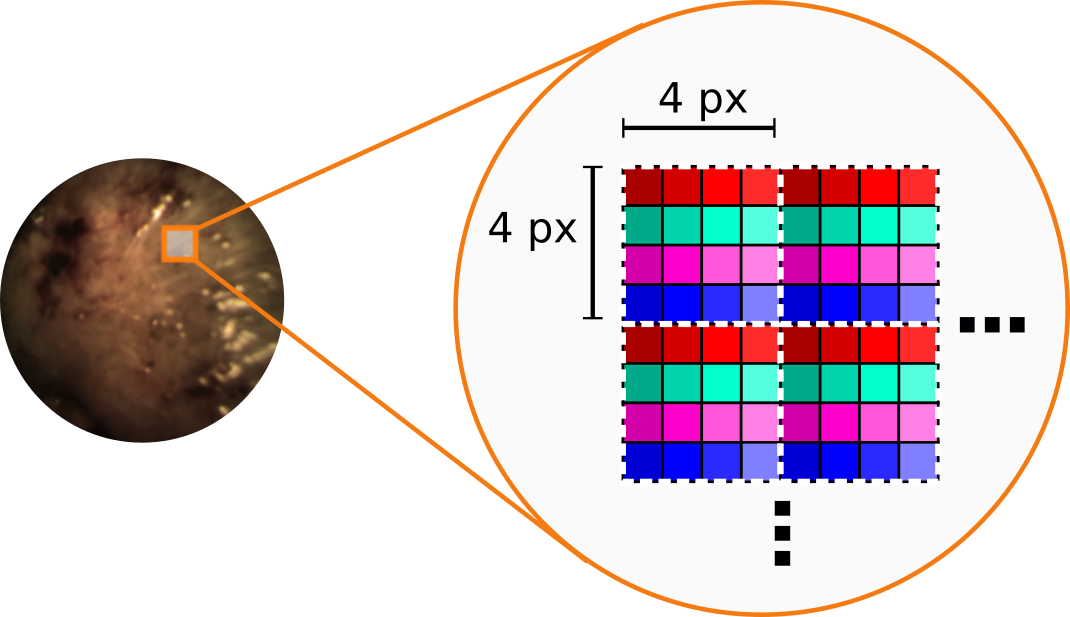}
    \caption{Representation of the 4x4 mosaic pattern of the multispectral camera sensor. Each colored square represents a different filter; these filters form a 4x4 pattern that extends over the whole image.}
    \label{fig:camera_mosaic}
\end{figure}

\begin{figure}[h!]
    \centering
    \includegraphics[width=0.9\textwidth]{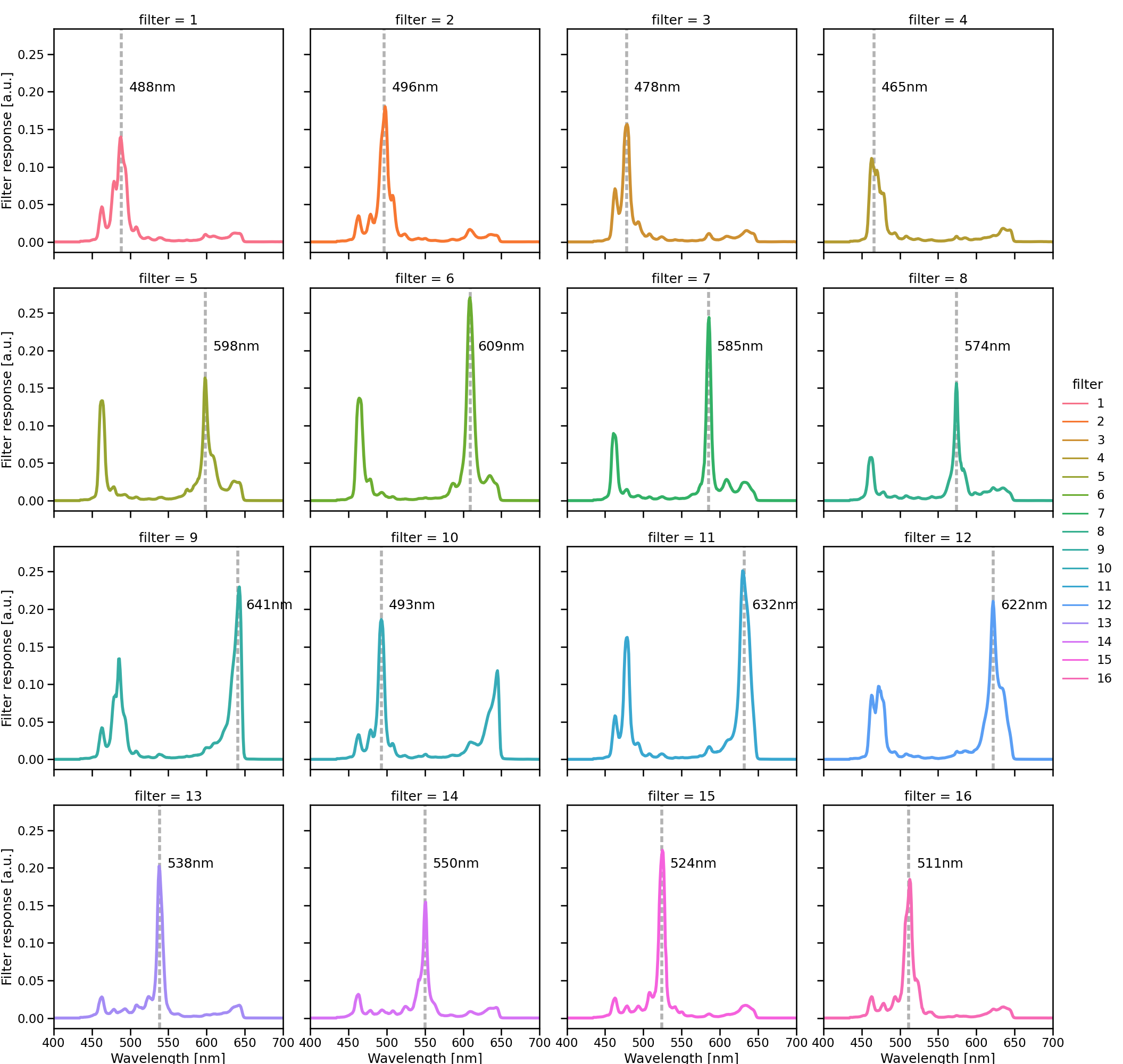}
    \caption{The filter responses of the 16 Ximea camera bands. Some bands, such as 5-12, show two extra peaks in the spectral response, which are referred to as ``second order'' peaks.}
    \label{fig:filter_responses}
\end{figure}

\paragraph{Multispectral camera} The core component is a xiQ multispectral camera (MQ022HG-IM-SM4x4, XIMEA GmbH, Muenster, Germany), which is a small (26x26x31mm) and light (32g) multispectral camera. It is based on the imec (Leuven, Belgium) mosaic snapshot sensor, which acquires 16 spectral bands at a single snapshot, using a 4x4 repeating mosaic pattern of filters (Fig. \ref{fig:camera_mosaic}); the spectral response of each filter is shown in Fig. \ref{fig:filter_responses}. Several of the bands show two peaks in the spectral response. These are caused by the measurement principle of the Fabry-P\'erot filters \cite{Poirson_97}, which lead to so-called ``second order'' peaks. The intensity of such peaks depends on the height of the filters' optical cavity, the refractive index of the sensor material and the cosine of the light incidence angle \cite{Koonen2006}.

\begin{figure}[h!]
    \centering
    \includegraphics[width=0.6\textwidth]{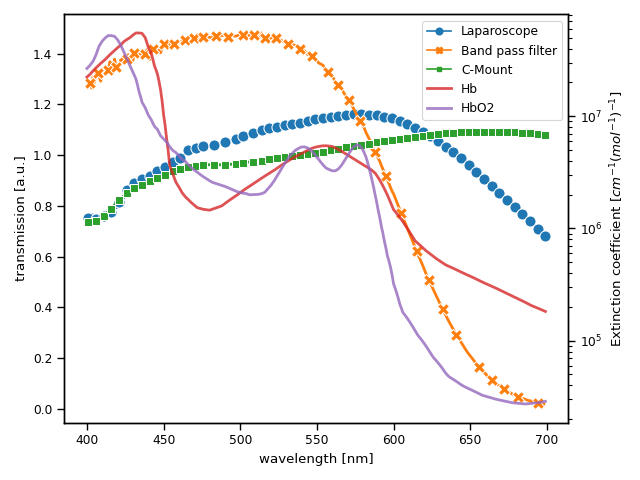}
    \caption{$\ell_1$ normalized transmission spectra of the laparoscope, band pass filter and C-Mount adapter are shown in the left axis. Extinction coefficients of oxyhemoglobin and deoxyhemoglobin are shown on the right axis. The band pass filter mainly filters light in the low wavelength region where blood absorbs the most (below 600nm), thus making the spectral power distribution that reaches the camera detector more uniform across wavelengths.}
    \label{fig:opt_components}
\end{figure}

\paragraph{Surgery-specific components} Due to intrinsic optical tissue properties, the reflectance of human tissue in the red region (above $\sim$ 620nm) is higher than the one in the blue region (below $\sim$ 490nm). To ensure more balanced camera counts, and thus similar noise levels across different camera filters, a 335 - 610nm  band pass filter (FGB37, Thorlabs Inc., Newton, Newton, New Jersey, United States) was placed between a C-Mount adapter and a laparoscope. The C-Mount adapter (20200043, KARL STORZ SE \& Co. KG, Tuttlingen, Germany) features an adjustable focal length, with a maximum of 38mm. To enable the recording of multispectral images during minimally-invasive surgery, the camera was connected to a standard $30^\circ$ laparoscope (26003BA, KARL STORZ SE \& Co. KG, Tuttlingen, Germany) via the aforementioned C-Mount adapter. The $\ell_1$ normalized transmission spectra of the band pass filter, the laparoscope and the C-Mount adapter are shown in Fig. \ref{fig:opt_components}. The extinction coefficients of oxyhemoglobin and deoxyhemoglobin are also shown as reference.  As light source, we chose a Xenon light source (IP20, Richard Wolf GmbH, Knittlingen, Germany) as it provides brighter and more uniformly distributed light intensity across different wavelengths in comparison to a Halogen light source. 

\paragraph{Image recording and storing} For image acquisition and recording, a standard laptop (Msi GE75 Raider 85G, intel i7, NVIDIA RTX 2080) was used. A custom C++ software (not publically available) based on the XIMEA Application Programm Interface (XiAPI, XIMEA, Muenster, Germany) was created to record the multispectral images.

\subsection{First-in-human application}
\label{subsec:in_vivo}

To demonstrate the capability of the proposed system for video-rate multispectral image acquisition of tissue in mininmally-invasive surgery, we performed the following proof-of-concept experiment. 

\paragraph{\textit{In vivo} image acquisition} Recordings were taken from a patient undergoing partial nephrectomy in collaboration with the Urology Department of the Staedtisches Klinikum Karlsruhe (Karlsruhe, Germany). The experiment was composed of two main phases: 
\begin{enumerate}
\item \textit{Before clamping:} Multispectral images of the surface of the kidney were recorded for 45s before the clamp was applied to the renal artery. This amounted to $\sim$1200 multispectral images.
\item \textit{During clamping:} The renal artery was clamped, and the kidney surface was monitored at video rate by the MSI system for 45s, amounting to an extra $\sim$1200 multispectral images. The success of the clamping procedure was confirmed by ICG injection into the patient's bloodstream and subsequent visualization of the fluorescence signal with the da Vinci Robot\textsuperscript{\textregistered} (Intuitive Surgical Deutschland GmbH, Freiburg, Germany). The ICG injection was prepared by mixing 50mg of ICG powder (PULSION Medical Systems SE, Feldkirchen, Germany) with 10ml of distilled water.
\end{enumerate}
Between the two recordings (before and during clamping), the multispectral laparoscope needed to be retracted, such that the surgical team could perform the clamping procedure through the trocar used for the laparoscope. Data analysis was performed as follows:
first one region of interest (ROI) was selected and subsequently tracked across consecutive frames, next, the high dimensional spectral data was normalized and visualized in two dimensions.

\paragraph{Region of interest tracking} One ROI was selected on each image sequence (before clamping and during clamping) based on the requirement that a) there was no visible fat on the tissue surface and b) none of the  multispectral bands were underexposed or overexposed within the ROI.  The ROI was defined in the first image and subsequently tracked across consecutive frames with a deep learning-based algorithm. As the multispectral laparoscope needed to be retracted between the two recordings (before and during clamping), we were not able to ensure imaging of the exact same region. Hence, the ROIs of the two phases may not correspond perfectly.
The tracking was performed on RGB images reconstructed from the multispectral images. The reconstructed RGB images were fed into a pre-trained VGG19 \cite{simonyan2015deep} neural network and deep features were extracted from its seventh convolutional layer. These extracted features were further processed by the tracker Discriminative Correlation Filter with Channel and Spatial Reliability (CSRDCF) \cite{Lukezic2018}. The ROIs generated with the CSRDCF tracker were visually inspected on each frame to ensure that there was no unwanted ROI location drift or sudden jumps. The image intensity on each tracked ROI was also analyzed to ensure that there were no underexposed or overexposed pixels. 

\paragraph{Data visualization} For 2D visualization of the high-dimensional spectral data, the data from each ROI was first normalized with a white ($W$) and dark($D$) reference recording taken with a Spectralon\textsuperscript{\textregistered} target (Edmund Optics, Barrington, USA). Given an ROI of dimensions $N\times M \times B$, where $(N,M)$ are the spatial dimensions and $B$ is the number of spectral bands, the intensity of each pixel at spatial location $(i,j)$ and spectral band $k$ was normalized according to:
\begin{equation}
  (\bar{I}_{(i,j)})_k = \frac{(I_{(i,j)})_k - (D_{(i,j)})_k}{(W_{(i,j)})_k - (D_{(i,j)})_k}  
  \label{eq:white_dark_norm}
\end{equation}
After normalization with the white and dark reference measurements, the median spectra within each ROI were computed. Subsequently, an $\ell_1$ normalization across different bands was performed to compensate for the influence of light source intensity changes due to changes in the distance of the laparoscope to the surface of the kidney. The resulting spectra  $(\hat{I}_{(i,j)})_k$ can be compared between different image sequences. 

Uniform Manifold Approximation and Projection (UMAP)\cite{McInnes2018} was chosen to reduce the number of dimensions of the spectra from 16 to 2 dimensions. The two-dimensional representation of the normalized spectra $(\hat{I}_{(i,j)})_k$ corresponding to the image sequences described in Sec. \ref{subsec:in_vivo} was optimized by performing a random search of the following parameters:  
\begin{itemize}
    \item Parameter 1: The effective minimum distance between embedded points. Random samples were uniformly selected in the range (0, 1).
    \item Parameter 2: The size of the local neighbourhood. Random integer values were uniformly sampled in the range [2, 100] as suggested by \cite{McInnes2018}.
\end{itemize}
A total of 1000 points were sampled from the space defined by parameter 1 and parameter 2, and one UMAP model was trained for each combination of parameters. The performance of each trained model was further evaluated by computing the Kullback-Leibler (KL) divergence \cite{Joyce2011} between the two sets of normalized spectra $(\hat{I}_{(i,j)})_k$ corresponding to the image sequences described in Sec. \ref{subsec:in_vivo}. The best parameters were then selected based on the models that minimize the KL divergence in addition to a visual inspection of the transformed spectra.
The optimal values for each of the parameters are $0.8$ for the effective minimum distance and $5$ for the size of the local neighbourhood.

\section{Results}

\begin{figure}[h!]
    \centering
    \includegraphics[width=0.6\textwidth]{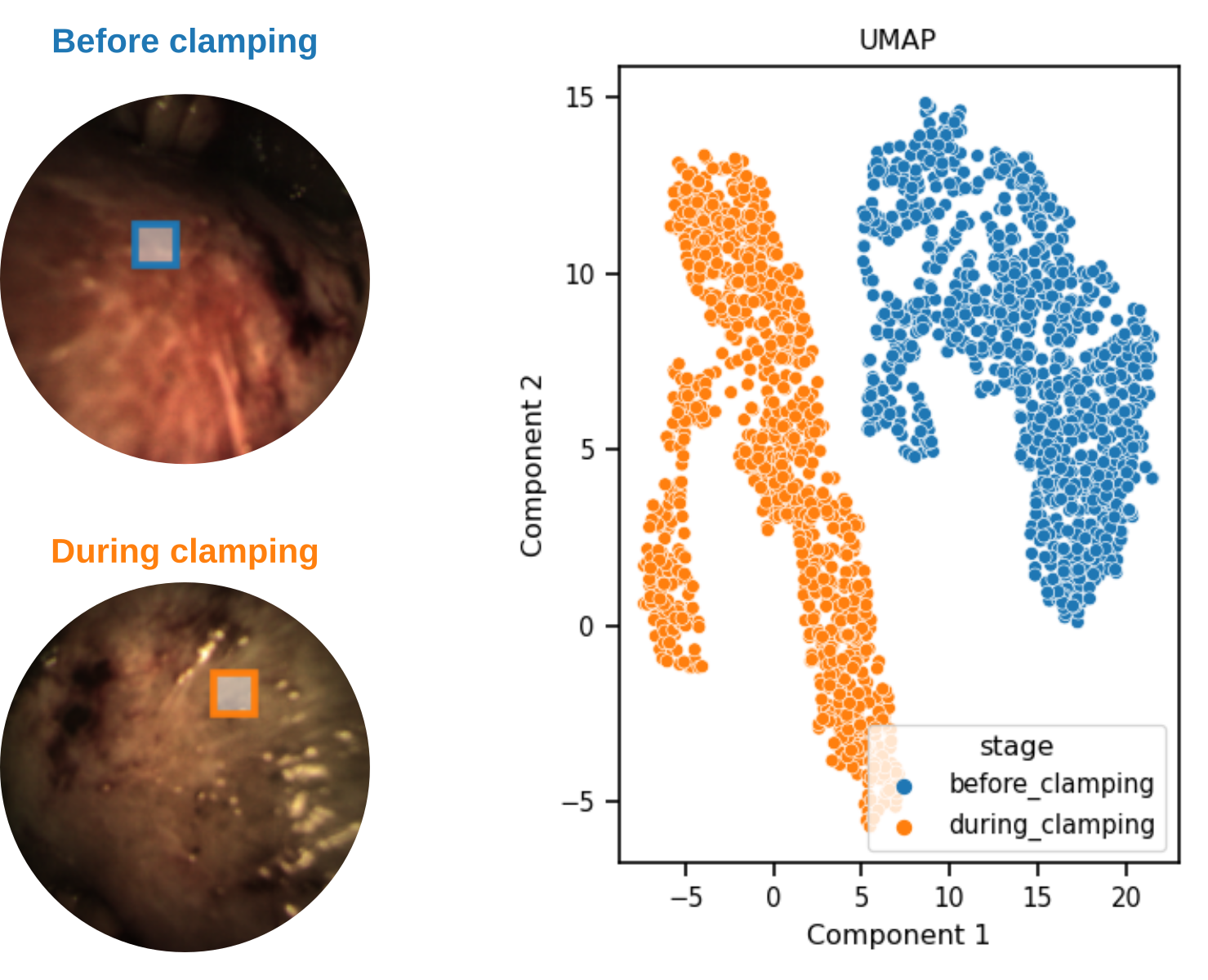}
    \caption{UMAP of spectra selected from one ROI before and during clamping. The displayed RGB images were reconstructed from the multispectral data. Each point in the UMAP represents one point in time and corresponds to the median spectrum in the ROI followed by an $\ell_1$ normalization in one frame ($\sim$1200 frames acquired in 45 sec).}
    \label{fig:umap}
\end{figure}

We successfully applied our multispectral imaging system \textit{in vivo}. 
The selected ROIs and two sample images of both sequences are shown in Fig. \ref{fig:umap}. A short video showing the ROI tracking results for each image sequence can be found \href{https://zenodo.org/record/4836161#.YLDnzyaxW1o}{online}.
The two-dimensional representation of the normalized spectra corresponding to each image sequence is shown in Fig. \ref{fig:umap}.

\section{Discussion}
\label{sec:live_discussion}
To our knowledge, we are the first to present a lightweight laparoscopic multispectral camera system capable of video-rate imaging in the operating room. Our proof-of-concept experiment indicates that the proposed system can monitor changes in kidney perfusion. In an ongoing study at the Urology Department of the Staedtisches Klinikum Karlsruhe (Karlsruhe, Germany), we are now testing the hypothesis that the more invasive ICG-based approach can be replaced by MSI for detecting ischemia during partial nephrectomies. Further studies could also determine if oxygenation estimations based on multispectral recordings \cite{Ayala_2019} can serve as an outcome predictor for postoperative renal function, as studied by Best et al. \cite{best_renal_2013}. If this were possible, the surgeon could adapt the surgery to the respective current state of the kidney, e.g. by early unclamping in the case of otherwise foreseeable long-term damage due to low oxygenation estimations.

The key strengths of our imaging system are the high acquisition speed (above 25 multispectral images per second), the low weight (32g) and the small size (camera cube with edge size of 26x26x31mm). However, the snapshot technique also comes with several limitations. The second order peaks of the multispectral camera filters shown in Fig. \ref{fig:filter_responses} can potentially cause challenges in image interpretation and analysis. For example, given the fact that the second order peaks are located at low wavelengths in the optical range, and that hemoglobin and deoxyhemoglobin absorve light mainly in those spectral regions, such second order peaks should be taken into account when the target application is perfusion estimation or oxygenation estimation. In spite of this challenge, we showed that the spectra corresponding to the before and during clamping image sequences of the kidney can be clustered with UMAP, as illustrated in Fig \ref{fig:umap}.  

In conclusion, we showed that video-rate multispectral imaging in the operating room is possible and offers the potential for real-time ischemia detection during surgery. Future work will be directed to fully-automatic image interpretation and systematic validation studies.

\subsection*{Acknowledgements}
This project has received funding from the European Union’s Horizon 2020 research and innovation program through the ERC starting Grant COMBIOSCOPY under Grant agreement No. ERC-2015-StG-37960 as well from the  Helmholtz Association under the joint research school “HIDSS4Health – Helmholtz Information and Data Science School for Health". We thank Minu Tizabi for proofreading.

\subsection*{Ethics}
All experiments involving humans were performed in accordance with the Declaration of Helsinki and all protocols were approved under the reference No. B-F-2019-101 by Landesaerztekammer Baden-Wuerttemberg (Germany).

\printbibliography

\end{document}